\pdfoutput=1
\documentclass[11pt]{article}

\usepackage[final]{acl}
\usepackage{multirow}

\usepackage{times}
\usepackage{latexsym}
\usepackage{array}
\usepackage{booktabs}
\usepackage{graphicx}
\usepackage{tcolorbox}
\usepackage{adjustbox}
\usepackage{tabularx}
\usepackage[T1]{fontenc}
\usepackage[utf8]{inputenc}

\usepackage{microtype}

\usepackage{inconsolata}

\usepackage{graphicx}

\title{Adapting Multilingual Embedding Models to Historical Luxembourgish}

\author{Andrianos Michail \\
    University of Zurich \\
    \texttt{andrianos.michail@cl.uzh.ch} \And
    Corina Julia Raclé \\
    University of Zurich \\
    \texttt{corinajulia.racle@uzh.ch}  \AND
    Juri Opitz \\
    University of Zurich \\
    \texttt{jurialexander.opitz@cl.uzh.ch}  \\ \And
    Simon Clematide \\
    University of Zurich \\
    \texttt{simon.clematide@cl.uzh.ch} \\ 
}

\begin{document}
\maketitle
\begin{abstract}
\begin{itemize}
The growing volume of digitized historical texts requires effective semantic search using text embeddings. However, pre-trained multilingual models face challenges with historical content due to OCR noise and outdated spellings. This study examines multilingual embeddings for cross-lingual semantic search in historical Luxembourgish (LB), a low-resource language. We collect historical Luxembourgish news articles from various periods and use GPT-4o for sentence segmentation and translation, generating 20,000 parallel training sentences per language pair. Additionally, we create a semantic search (Historical LB Bitext Mining) evaluation set and find that existing models perform poorly on cross-lingual search for historical Luxembourgish.  Using our historical and additional modern parallel training data, we adapt several multilingual embedding models through contrastive learning or knowledge distillation and increase accuracy significantly for all models. We release our adapted models and historical Luxembourgish-German/French/English bitexts to support further research. \footnote{See \href{https://github.com/impresso/histlux_emb}{https://github.com/impresso/histlux\_emb} for our released models, data and source code.}
\end{itemize}
\end{abstract}

\section{Introduction}

Exploration possibilities of historical texts, such as newspapers, have advanced rapidly due to digitization efforts by libraries and archives \citep{EhrmannBunoutClavert+2023+1+22}. Traditionally, tools relied on keyword-based searches, often enhanced with semantic enrichment techniques such as named entity recognition \citep{EhrmannHamdi:2023}. 

Recent embedding benchmarks \citep{muennighoff-etal-2023-mteb, enevoldsen2025mmtebmassivemultilingualtext} show that massively multilingual embedding models, trained on diverse multilingual corpora, perform well in both multilingual and cross-lingual semantic search. These models have also become integral in Retrieval-Augmented Generation (RAG), where they help retrieve more relevant and contextually appropriate documents, thereby improving the faithfulness of generated responses.

However, for low-resource languages like Luxembourgish (LB), where multilingual models have limited exposure, their performance remains uncertain. Applying these models to semantic search in imperfectly digitized historical collections introduces additional challenges, as they must handle OCR errors and historical spelling variations. The disparity between these noisy, historical texts and the clean, modern digital-born data used to train multilingual models, combined with their limited support for Luxembourgish, complicates the development of effective exploration tools for historical Luxembourgish newspaper archives. 

To address this issue, we compile 2,338 historical Luxembourgish news articles from different time periods and use GPT-4o to segment and translate them into modern French (FR), English (EN) and German (DE). The resulting parallel sentences serve as fine-tuning data to adapt existing multilingual embedding models for imperfectly digitized historical Luxembourgish.

Our main contributions: \newline
(1) We adapt multilingual embeddings for digitized historical Luxembourgish by generating training data through a prompt-based translation approach with GPT-4o.  
\newline
(2) We define a historical bitext mining task and create a high-quality cross-lingual semantic search test set with 233 source news articles (LB-DE: 2,127; LB-FR: 2,157; LB-EN: 2,105 sentences).
\newline
(3) We fine-tune and evaluate off-the-shelf models -- \textit{M-MPNet} \citep{reimers-gurevych-2020-making}, \textit{LaBSE} \citep{feng-etal-2022-language}, \textit{M-GTE} \citep{zhang-etal-2024-mgte}, and \textit{LuxEmbedder} \citep{philippy-etal-2025-luxembedder} -- to assess our adaptation methods.  
\newline
(4) We propose and evaluate a 1:1 data mixing strategy that balances noisy historical texts with clean modern texts to minimize performance degradation on modern Luxembourgish benchmarks.

\section{Related Work}

This section reviews relevant embedding models that support Luxembourgish semantic search, including monolingual Luxembourgish models and multilingual embeddings.

\citet{reimers-gurevych-2020-making} use knowledge distillation through a strong paraphrase-trained English embedding model and parallel data to create cross-lingually aligned models. Multiple instances of such models have been open sourced and a particularly powerful and popular one is \textit{paraphrase-multilingual-mpnet-base-v2} (\textbf{M-MPNet}) which was trained on over 50 languages. Later within this work, we will explain how we extend this model to also support Luxembourgish.

The multilingual bitext mining model \textbf{LaBSE} \citep{feng-etal-2022-language} is trained with translation ranking loss and negative samples.  It has been trained roughly on less than 100 Luxembourgish-English sentence pairs and specializes in zero-shot bitext mining. 

A recent model is GTE Multilingual (\textbf{M-GTE}) \citep{zhang-etal-2024-mgte}, a multilingual embedding model designed for long context text representation and reranking.  \textit{M-GTE} has been trained with hard negatives and has included 50,000 Luxembourgish pairs within its contrastive pre-training.

Specific model adaptations to Luxembourgish have also been developed. One example is \textbf{LuxemBERT} \citep{lothritz-etal-2022-luxembert}, a monolingual BERT model pre-trained for Luxembourgish using augmented data, partially generated by translating texts from closely related languages and incorporating relevant text sources. 

Closely related to our work, \textbf{LuxEmbedder} \citep{philippy-etal-2025-luxembedder} used OpenAI's \textit{text-embedding-3-small} and \textit{LaBSE} to mine a set of parallel sentences for each pair of languages between Luxembourgish, English, and French. These parallel sentences (up to 20,000 per pair) were then used to further fine-tune \textit{LaBSE}, improving performance on modern Luxembourgish evaluation sets. However, its ability to handle Luxembourgish texts from different historical periods---potentially affected by digitization errors common in large-scale historical text collections, remains unclear.

Our work aims to extend existing embedding models to better perform cross-lingual semantic search within a collection of historical, OCR-noisy Luxembourgish texts. The conditions of these texts combined with the different spelling variations\footnote{Luxembourgish had no standardized spelling until 1946 and underwent multiple further reformations (eg. in 1999)} poses an interesting generalization challenge to the models.

\section{Method}

To adapt and evaluate embedding models for digitized historical Luxembourgish news articles, we create parallel texts by translating them into modern German, French, and English. This allows the models to learn cross-lingual representations and improves their ability to align historical Luxembourgish with contemporary languages for semantic search.

\subsection{Parallel Historical Luxembourgish}

We build our translated parallel data sets LB-DE, LB-FR and LB-EN from monolingual Luxembourgish texts sourced from the publicly available BNL newspaper archive.\footnote{\href{https://data.bnl.lu/data/historical-newspapers/}{https://data.bnl.lu/data/historical-newspapers}} Our data consists of articles from newspapers published between 1841 and 1948. To select diverse samples for translation,  we first cluster the articles into 2,000 groups by K-Mean on a 100-topics LDA model output\footnote{Taken from the \href{https://impresso-project.ch}{impresso-project.ch} \citep{ehrmann-etal-2020-language}.} and keep the 605 clusters with more than 20 articles. 

We select articles through a two-step process, resulting in a total of 2,340 articles, as shown in Figure 1. First, we retrieve the most representative article from each cluster, ensuring it contains between 5 and 20 sentences (cutting of the remaining sentences). In a second round, we randomly sample three additional articles per cluster under the same length conditions.

\begin{figure}[t]
    \centering
    \includegraphics[width=\linewidth]{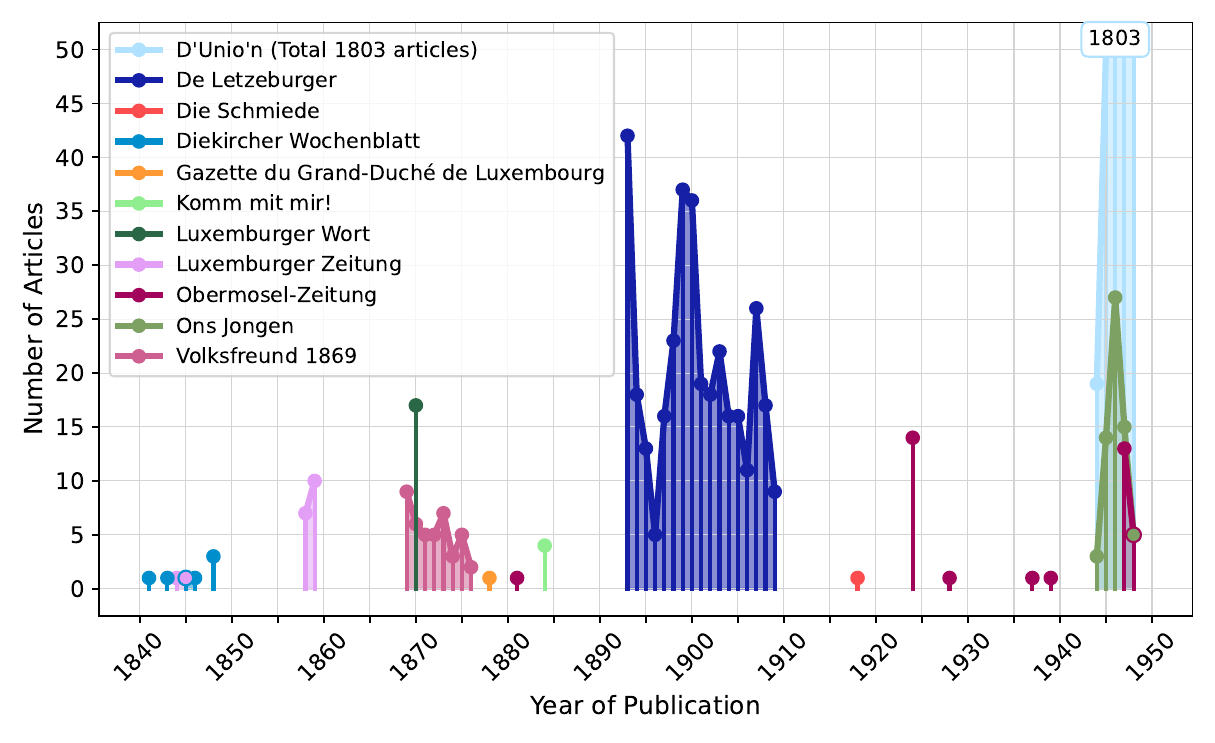} 
    \caption{Source LB articles per newspaper per year.}
    \label{fig:right-column}
\end{figure}

We prompt GPT-4o to segment historical Luxembourgish articles and generate sentence-level translation pairs separately for German, French and English (see Prompt \ref{fig:prompt_step1}). The model is instructed to preserve the original meaning and structure as closely as possible while reconstructing sentences affected by OCR errors that could hinder translation. This process yields approximately 22,500 sentence pairs for the LB-DE, LB-FR and LB-EN pair. Notably, GPT-4o appears to perform sentence segmentation consistently resulting in 65.0\% of sentences forming exact quadruplets (4-way parallel) across the four languages. 

To ensure fidelity to the original articles, we calculate the percentage of regenerated Luxembourgish sentences that do not exactly match their source texts. These account for 1.4\% of all historical Luxembourgish sentences per language pair, which we manually correct. Most mismatches result from missing or added punctuation, modernized spelling, and, in rare cases, errors caused by the LLM not adhering to the instructed format.
% Exact numbers LUX-FR: 358/22’369 = 1.6%
% LUX-DE: 301/22’397 = 1.34%
% LB-EN: 1.36%
% Average 1.43%

% Newer version, older draft in comments.

To assess translation quality, a quadrilingual native speaker of Luxembourgish (LB) annotated 100 randomly selected sentence quadruplets after removing 15 samples with severe OCR problems. Of the 100 LB sentences presented without context, 88 were judged to be comprehensible or at least confidently guessable (23). The remaining 12 were considered incomprehensible due to OCR errors and archaic spellings, and their translations were not evaluated.

For the comprehensible and confidently guessable sentences, the German translations were rated as adequate in 78 cases (88.6\%), with missing minor details in 9 cases and 1 case of inadequate translation. The French translations showed a similar pattern: 78 were adequate (88.6\%), 9 were missing minor details, and 1 was inadequate. The English translations were also adequate in 78 cases (88.6\%), with 10 missing minor details. A sample of the annotated dataset is available in the appendix (Table~\ref{tab:human_annotation}).

\subsection{Framing an Evaluation Task: Historical LB Bitext Mining}
From our parallel dataset, we set aside a held-out test set of 233 articles (2,127 sentences) to establish a historical semantic search benchmark for Luxembourgish-to-German, French, and English bitext mining (LB<->DE/FR/EN). A prediction is considered a true positive if the embedding model assigns a higher similarity to the correct parallel sentence than to any of the 2k alternative candidates. We report the bidirectional average accuracy. To minimize false negatives caused by near-identical sentences, we exclude candidate sentences with a Levenshtein similarity score above 0.85 to the source sentence, after removing non-alphanumeric characters from both. This filtering affects 57 source-candidate pairs (2.7\%) in German, 65 (3\%) in French, and 76  (3.6\%) in English. A human review at different thresholds confirms the appropriateness of the filtering process and the chosen threshold.

\subsection{Modern LB Evaluation Tasks} 

We replicate two evaluation tasks on modern Luxembourgish from \citet{philippy-etal-2025-luxembedder}.

\textbf{ParaLux} is a monolingual paraphrase detection test set designed to evaluate embedding models. Performance is measured by the proportion of cases (a total of 312 triplets) in which an embedding model assigns a higher similarity to an anchor-positive pair than to an anchor-negative pair.  The negative sentences are adversarially generated to maintain high lexical similarity and manually verified to ensure they are true negatives.

\textbf{SIB 200 (LB)} is a repurposed subset of the `Flores' dataset \citep{costa2022no, adelani-etal-2024-sib}, used for monolingual zero-shot topic classification. In this task, texts are assigned to template sentences representing candidate topics based on embedding similarity. 

\subsection{Adapting Multilingual Embedding Models to Historical LB}

\subsubsection{Datasets}

\begin{table*}[t]
\centering
\renewcommand{\arraystretch}{1.0}
\adjustbox{width=\linewidth}{\begin{tabular}{ll|rrrr|rrr}
\toprule
 \multirow{2}{*}{\textbf{Model}} &  \multirow{2}{*}{\textbf{Training Data}} & \multicolumn{4}{c|}{\textbf{Historical LB Bitext Mining}} & \multicolumn{3}{c}{\textbf{Modern LB}} \\
 \cmidrule(lr){3-6} \cmidrule(lr){7-9}
 & & \textbf{LB$\leftrightarrow$FR} & \textbf{LB$\leftrightarrow$EN} & \textbf{LB$\leftrightarrow$DE} & \textbf{AVG} & \textbf{SIB 200 (LB)} & \textbf{ParaLux} & \textbf{AVG} \\
\midrule
Random Baseline & -- & 00.00 & 00.00 & 00.00 & 45.97 & 14.28 & 50.00 & 32.14 \\
\raisebox{-0.28\height}{\includegraphics[width=0.55cm]{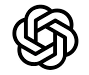}}text-embedding-3-small & -- & 78.36 & 75.08 & 82.33 & 78.59 & 40.20 & 15.71 & 27.96 \\
\raisebox{-0.28\height}{\includegraphics[width=0.55cm]{figures/openailogo.png}}text-embedding-3-large & -- & 86.18 & 83.63 & 88.15 & 85.99 & 58.82 & 26.28 & 42.55 \\
\midrule
\textit{M-MPNet} & -- & 46.32 & 45.04 & 46.55 & 45.97 & 24.71 & 26.60 & 25.66 \\
M-MPNet(+LB Distilled)  & LB$\leftrightarrow$EN (Hist) & 87.23 & 87.53 & 89.14 & 87.97 & 42.65 & 56.09 & 49.37 \\ 
& LB$\leftrightarrow$EN (Modern) & 75.55 & 77.03 & 78.09 & 76.89 & \textbf{59.51} & 80.13 & 69.82 \\
& LB$\leftrightarrow$EN (Mixed) & \textbf{89.32} & \textbf{89.55} & \textbf{91.44} & \textbf{88.79} & 59.41 & \textbf{80.45} & \textbf{70.48} \\
\midrule
\textit{LaBSE} & -- & 93.12 & 95.27 & 94.01 & 94.13 & 43.24 & \textbf{38.14} & 40.69 \\
LaBSE (Hist) & LB$\leftrightarrow$FR & \textbf{97.73} & 97.22 & 98.10 & 97.68 & 39.61 & 25.00 & 32.31 \\
 & LB$\leftrightarrow$EN & 97.24 & 97.44 & 97.96 & 97.54 & 41.76 & 22.44 & 32.10 \\
 & LB$\leftrightarrow$DE & 97.08 & 97.01 & \textbf{98.52} & 97.54 & 34.02 & 14.74  &  24.38 \\
LaBSE (Mixed)  & LB$\leftrightarrow$FR & 97.40 & \textbf{97.55} & 98.22 & 97.35 & \textbf{45.69} & 31.73 & 47.66 \\
 & LB$\leftrightarrow$EN & 96.80 & 97.34 & 97.82 & \textbf{97.75} & 45.59 & 36.86 & \textbf{50.23} \\
\midrule
\textit{LuxEmbedder} & -- & 84.49 & 85.09 & 85.48 & 85.02 & \textbf{65.59} & \textbf{52.24} & \textbf{58.92} \\
LuxEmbedder (Hist) & LB$\leftrightarrow$FR  & \textbf{97.47} & 97.51 & 98.24 & 97.74 & 50.39 & 32.37  & 41.38 \\
 & LB$\leftrightarrow$EN  & 97.18 & 97.29 & 98.26 & 97.58 & 54.12 & 28.85  & 41.49 \\
 & LB$\leftrightarrow$DE  & 97.25 & \textbf{97.72} & \textbf{98.43} & \textbf{97.80} & 46.76 & 26.60  & 36.68 \\
LuxEmbedder (Mixed) & LB$\leftrightarrow$FR  & 96.97 & 97.32 & 97.77 & 97.72 & 56.86 & 38.46  & 38.71 \\
 & LB$\leftrightarrow$EN  & 97.41 &97.58 & 98.26 & 97.32 & 56.86 & 43.59  & 41.23 \\
\midrule
\textit{M-GTE} & -- & 83.68 & 80.12 & 87.55 & 83.78 & 55.78 & \textbf{70.51} & 63.20 \\
M-GTE (Hist) & LB$\leftrightarrow$FR & 95.18 & 94.23 & 96.05  & 95.15 & 59.12 & 57.05  & 58.09 \\
 & LB$\leftrightarrow$EN & \textbf{95.81} & 95.56 & 96.52 & \textbf{95.96} & 54.71 & 55.77  & 55.24 \\
 & LB$\leftrightarrow$DE & 95.23 &94.61 & \textbf{97.65} & 95.83 & 45.29	& 42.31 & 43.80 \\
M-GTE (Mixed) & LB$\leftrightarrow$FR  & 95.53 & 95.11 & 96.78 & 95.80 & 60.98 & 60.26  & 60.62 \\
 & LB$\leftrightarrow$EN  & 95.48 & \textbf{95.58} & 96.55 & 95.87 & \textbf{67.84} & 64.10 & \textbf{65.70} \\
\midrule
M-GTE (Hist, Modern: 120k) & LB$\leftrightarrow$DE/FR/EN  & \textbf{96.83} & \textbf{97.15} & \textbf{97.93} & \textbf{97.30} & 62.16 & 62.82  & 62.75 \\
\bottomrule
\end{tabular}}
\caption{Performance (accuracy) of the examined models and our adapted variants within the Historical and Modern Luxembourgish evaluation sets. The last row shows an adapted model trained on the maximum available data, with details found at the end of Section~\ref{sec:final_mix}.}
\label{tab:evaluation_results}
\end{table*}

\textit{Hist}orical: We use 2,105 historical LB newspaper articles (excluding held-out articles) with their sentence-level translations to create a parallel training set for the following language pairs: LB-DE (20,092), LB-FR (20,010), and LB-EN (19,054) sentences.

\textit{Modern}: \citet{philippy-etal-2025-luxembedder} extracted 89,405 LB-FR and 28,172 LB-EN parallel sentence pairs  from \href{https://www.rtl.lu/}{RTL.lu}, a trilingual news platform. This dataset was used to fine-tune the \textit{LuxEmbedder} model.

\paragraph{Training Data Configurations}  
We investigate three data mixing strategies for model training:  \newline
(1) \textit{Hist}orical:  20,000 translated sentence pairs (LB$\leftrightarrow$FR, LB$\leftrightarrow$EN, or LB$\leftrightarrow$DE) from historical texts.\newline  
(2) \textit{Modern}: 20,000 bitext-mined sentence pairs (LB$\leftrightarrow$FR or LB$\leftrightarrow$EN) from modern Luxembourgish news. \newline
(3) \textit{Mixed}: 20,000 \textit{Hist} sentence pairs  with 20,000 \textit{Modern} sentence pairs in mixed batches.

% These data configurations serve as the foundation for the fine-tuning strategies in the subsequent sections.

\subsubsection{LB Knowledge Distillation}

We adapt \textit{M-MPNet} for historical LB using multilingual knowledge distillation \citep{reimers-gurevych-2020-making}. The original English model \textit{paraphrase-mpnet-base-v2} serves as a frozen teacher, while LB-EN parallel sentences are used to train the \textit{M-MPNet} student model to embedded LB sentences similar to their English translations.
We fine-tune \textit{M-MPNet} for five epochs using each of the three data mixing strategies: (1) \textit{Hist}orical, (2) \textit{Modern}, and (3) \textit{Mixed}.

\subsubsection{Contrastive Loss}

We adapt \textit{LaBSE}, \textit{LuxEmbedder}, and \textit{M-GTE} to historical LB using contrastive learning. Specifically, we fine-tune these embedding models using \textit{MultipleNegativesRankingLoss} \citep{henderson2017efficientnaturallanguageresponse}, with a batch size of 8 for one epoch.
For fine-tuning, we apply two of the previously defined data mixing strategies: (1) \textit{Hist}orical and (3) \textit{Mixed}.

\section{Results}

Table~\ref{tab:evaluation_results} shows the performance of the off-the-shelf and adapted models on the historical Luxembourgish bitext mining and the modern LB evaluation tasks.

Among the off-the-shelf(in cursive) models, \textit{LaBSE} is the strongest model in all three languages. 
Surprisingly, \textit{LuxEmbedder}, a LB-tuned version of \textit{LaBSE}, shows an average performance drop of 9pp across language pairs in our bitext mining task, despite improved performance on the modern LB tasks. Similarly, \textit{M-GTE} underperforms \textit{LaBSE} by 10.4pp. Both OpenAI embedding models (\textit{text-embedding-3-small/large}) show moderate performance. 

Among the models contrastively adapted using the \textit{Hist} pairs, the performance in historical bitext mining improves significantly, reaching up to 97.8\% accuracy. Notably, after domain adaptation, \textit{LuxEmbedder} matches the adapted \textit{LaBSE}, reaching over 97.8\% accuracy and closing the performance gap observed in the standard models. Meanwhile, the customized \textit{M-GTE} models lag behind by about 2pp. Interestingly, across all model architectures, training on any language pair improves performance similarly across all pairs, showing a positive cross-lingual transfer.

These models experience significant performance drops in modern LB evaluations, particularly in \textit{ParaLux}. However, adapting these models with mixed batches of \textit{Hist} and \textit{Modern} sentence pairs partially mitigated performance loss on the Modern LB evaluation tasks. Within \textit{LaBSE} and \textit{M-GTE}, this adaptation even improved the performance on SIB-200 topic classification, while sacrificing only up to 1\% of the performance on historical bitext mining. These mixed-data adapted models provide an overall stronger general backbone for cross-lingual semantic searching within collections containing both historical and modern Luxembourgish.

The \textit{M-MPNet} model, before distilling EN-LB knowledge, performs poorly in all LB evaluations, despite its proven exact matching capabilities in other languages, confirming its lack of support for the language. After distilling LB with any dataset, the model performs magnitudes better across the board. When distilled with a single dataset, the model performs best on historical semantic search tasks with the \textit{Hist} sentence pairs. In contrast, when distilled with \textit{Modern} sentence pairs, the model excels on modern LB tasks, achieving 80\% accuracy on ParaLux\footnote{As shown in recent work \citep{michail-etal-2025-paraphrasus}, results on adversarial paraphrase discrimination test sets might not accurately reflect performance on semantic search in general.  Therefore, this result should be interpreted with caution.} and outperforming the second-best \textit{M-GTE}, which achieves 70\%. Finally, distilling with the mixed data set yields the best results in all evaluations, demonstrating the synergy between the two sources.

However, even the \textit{Mixed}-data distilled \textit{M-MPNet} model only achieves an average accuracy across pairs of 90\% in historical bitext mining, trailing the contrastive domain-adapted \textit{LaBSE} and \textit{LuxEmbedder} by 8pp and the off-the-shelf \textit{LaBSE} model by 4pp.

\label{sec:final_mix}

\paragraph{The Final Model: Mix it All}  
For a final all-purpose model covering both historical and modern LB, we contrastively adapt \textit{M-GTE} to all language pairs of \textit{Hist} while preserving an equal number of \textit{Modern} sentence pairs, regardless of language. The adaptation dataset consists of 20,000 LB-FR/EN/DE (\textit{Hist}), 20,000 LB-EN (\textit{Modern}), and 40,000 LB-FR (\textit{Modern}), for a total of 120,000 sentence pairs.  

It is the best-performing historical semantic search \textit{M-GTE} model, achieving an average accuracy of 97.5\% across all language pairs. This model outperforms the adapted \textit{LaBSE} and \textit{LuxEmbedder} models on SIB-200 (+6pp) and ParaLux (+20pp), while performing similarly to them in the historical bitext mining evaluations.

\section{Conclusions}  

In this work, we explore the adaptation of multilingual embedding models to digitized historical LB texts, a task where off-the-shelf models struggle due to limited exposure and their reliance on clean modern data. To address this issue, we generate parallel sentence-segmented documents by translating historical Luxembourgish newspaper articles into French, English, and German using GPT-4o.  

To evaluate the effectiveness of adaptation, we design a historical bitext mining task with a held-out test set of 233 articles. Our results show that adaptation to parallel historical data improves retrieval accuracy by up to 13pp. However, this adaptation introduces trade-offs, particularly reducing performance on modern LB tasks that require high semantic precision, such as adversarial paraphrase detection. We mitigate this problem through a balanced data mixing strategy that helps preserve modern LB performance while improving historical text semantic search capabilities.

These results demonstrate the effectiveness of domain adaptation for historical text processing and suggest that such approaches could benefit low-resource languages facing digitization challenges. Such improvements are particularly relevant for libraries and archives, where effective cross-lingual semantic search can improve the discoverability of historical documents and support digital exploration.

\section*{Limitations}

Our findings paves the way for better semantic search systems within Luxembourgish archives. On the one hand, our method demonstrates clear benefits for the targeted use case, effectively embedding heterogeneous digitized historical texts and revealing shortcomings in off-the-shelf models. Through our exploration of adaptation methodologies, we have produced practical embeddings for semantic search while mostly preserving modern LB performance. On the other hand, we have not strictly reached a single best model across all evaluation sets. For example, in all of our adapted models, performance on \textit{ParaLux} declines, possibly indicating interference with modern LB understanding and reduced sensitivity to semantic nuances. 

Overall, we have applied a single adaptation method for each model type across all available data mixes, ensuring alignment with the models’ initial training methods. Exploring alternative adaptation approaches may reveal additional patterns.

One problem with our evaluation is that they are all at the sentence level, whereas applications of such models would often be at the paragraph, article, or document level. The hypothesis that our improved performance would be reflected when embedding longer segments of text is possible, but not guaranteed.
Lastly, while our research focuses on historical Luxembourgish, our methodology may also be useful for developing semantic search models in other underrepresented languages, which we do not examine in this study.

\section*{Acknowledgments}
We would like to thank Fred Phillipy for helping with the human annotation. This research is conducted under the project \textit{Impresso -- Media Monitoring of the Past II Beyond Borders: Connecting Historical Newspapers and Radio}. Impresso is a research project funded by the Swiss National Science Foundation (SNSF 213585) and the Luxembourg National Research Fund (17498891).
\bibliography{acl_latex}

\newpage

\appendix
\section{Appendix}
\label{sec:appendix}

\begin{figure}[h] \footnotesize
\begin{tcolorbox}[title=\small{\it{System:}},label={prompt},colback=white]
 You are a professional translator specializing in the translation of historical Luxembourgish newspaper articles into modern Standard \{German/French/English\}.
 
Your task is to translate paragraphs from such newspapers, provided to you by the user. These paragraphs may contain old spellings, outdated expressions, and likely a lot of OCR errors, as they are extracted from 19th-century LB newspapers. Please translate each sentence individually into modern Standard \{German/French\}. Prioritize retaining the original meaning, expressions, and any nuanced tone in each translation, even if the result sounds somewhat unconventional or even bad in \{German/French/English\}. If an expression is ambiguous due to its historical nature or OCR errors, attempt to reconstruct the most probable meaning based on linguistic context. Ensure that all punctuation and whitespace is preserved exactly. Do not add any extra formatting such as backticks, markdown, or additional symbols.\newline

Please return the source sentences and your translations in the following format as JSON:\newline
\texttt{\{"translation": [\newline
\{\texttt{"lb": "lb\_sent1", "\{de\}": "\{de\}\_sent1"}\},\newline
\{\texttt{"lb": "lb\_sent2", "\{de\}": "\{de\}\_sent2"}\},\newline 
\{\texttt{"lb": "lb\_sent3", "\{de\}": "\{de\}\_sent3"}\}, ...]}\}
\end{tcolorbox}

\caption{Zero-shot prompt template given to GPT-4o for the segmentation and translation of historical Luxembourgish newspaper articles to modern French(fr)/English(en)/German(de).}
\label{fig:prompt_step1}
\end{figure}

\begin{table*}[t!]
    \centering
    \renewcommand{\arraystretch}{1.2} % Adjust row height for compactness
    \small % Reduce font size for better fit
    \begin{tabular}{|p{1.3cm}|p{0.48cm}|p{6cm}|p{2.05cm}|p{1.5cm}|p{1.5cm}|p{1.5cm}|}
        \hline
        \textbf{Newspaper} & \textbf{Year} & \textbf{Sentence} & \textbf{LB Compr.} & \textbf{FR Compr.} & \textbf{EN Compr.} & \textbf{DE Compr.} \\
        \hline
        D'Union & 1946 & 
        \textbf{LB}: Si bestét aus 18 000 tuben, weit 30 tonnen a kascht 400 000 dollar. \newline
        \textbf{FR}: Elle est composée de 18 000 tubes, pèse 30 tonnes et coûte 400 000 dollars. \newline
        \textbf{EN}: It consists of 18,000 tubes, weighs 30 tons, and costs 400,000 dollars. \newline
        \textbf{DE}: Sie besteht aus 18.000 Röhren, wiegt 30 Tonnen und kostet 400.000 Dollar. 
        & Comprehensible & Adequate & Adequate & Adequate \\
        \hline
        Ons Jorgen & 1946 & 
        \textbf{LB}: „Daf net, mais ech hu kc properen Teller me'," \newline
        \textbf{FR}: "Certainement pas, mais je n'ai plus d'assiette propre," \newline
        \textbf{EN}: "Not really, but I don't have a single clean plate anymore." \newline
        \textbf{DE}: „Das nicht, aber ich habe keinen sauberen Teller mehr. 
        & Comprehensible & Adequate & Adequate & Adequate\\
        \hline
        De Letzeburger & 1893 & 
        \textbf{LB}: De Batti: Elo hätte mer d'Stemmung gut eriwer, hätte mer elo och nach Rén. \newline
        \textbf{FR}: Le Batti : Maintenant, nous aurions bien passé l'ambiance, si seulement nous avions aussi encore Rén. \newline
        \textbf{EN}: Batti: Now we would have a good atmosphere if we also had some rain. \newline
                \textbf{DE}: Der Batti: Jetzt hätten wir die Stimmung gut geschafft, hätten wir jetzt auch noch Regen.
        & Confidently Guessable & Adequate & Adequate & Adequate \\
        \hline
        De Letzeburger & 1905 & 
        \textbf{LB}: Op d'Weis: Das ist im Lehen hfisslich eingerichtet. \newline
        \textbf{FR}: À la manière de : Cela est arrangé vilaine dans la vie. \newline
        \textbf{EN}: To the tune: It is poorly arranged in life. \newline
        \textbf{DE}: Zur Melodie: Das ist im Leben hässlich eingerichtet.
        & Confidently Guessable & Adequate & Adequate & Adequate \\
        \hline
        De Letzeburger & 1893 & 
        \textbf{LB}: Wann och d'Liss'ché wéss ze feischtren. \newline
        \textbf{FR}: Même si Liss'ché sait lutter. \newline
        \textbf{EN}: Even if Lisette knows how to flirt. \newline
        \textbf{DE}: Wenn auch die Liss'ché weiß zu feilschen. 
        & Incomprehensible & / & / & /\\
        \hline
        Obermosel-Zeitung & 1924 & 
        \textbf{LB}: In vielen vorkern Bincl alle Krank, 80 cla,BB «lie ?eläer nicdt deBtellt terrien Können. \newline
        \textbf{FR}: Dans de nombreux villages, tous sont malades, si bien que les champs ne peuvent pas être cultivés. \newline
        \textbf{EN}: In many places, all are sick, so that the fields cannot be tended. \newline
        \textbf{DE}: In vielen Dörfern sind alle krank, so dass die Felder nicht bestellt werden können. 
        & Incomprehensible & / & / & / \\
        \hline
    \end{tabular}
    \caption{Sample of quadruplets of parallel sentence as used within our human evaluation of the dataset quality.}
    \label{tab:human_annotation}
\end{table*}

\end{document}